%% file: main.tex
\documentclass[10pt,twocolumn,letterpaper]{article}

\usepackage[pagenumbers]{iccv} 


\definecolor{iccvblue}{rgb}{0.21,0.49,0.74}
\usepackage[pagebackref,breaklinks,colorlinks,allcolors=iccvblue]{hyperref}

\usepackage{multirow}
\usepackage{enumitem}
\usepackage{booktabs}
\usepackage{tabularx}
\usepackage{graphicx}

\usepackage{ulem}
\usepackage{marvosym}

\newcommand\blfootnote[1]{%
  \begingroup
  \renewcommand\thefootnote{}\footnote{#1}%
  \addtocounter{footnote}{-1}%
  \endgroup
}

\renewcommand{\vec}[1]{\boldsymbol{#1}}
\renewcommand{\ie}{\textit{i.e.}}
\renewcommand{\eg}{\textit{e.g.}}
\newcommand{\mypara}[1]{\noindent\textbf{#1}}

\def\paperID{7038} 
\def\confName{ICCV}
\def\confYear{2025}

\title{Enhancing Spatial Reasoning in Multimodal Large Language Models through Reasoning-based Segmentation}

\author{
Zhenhua Ning$^{1,2,3{\dag}*}$ ~~~~Zhuotao Tian$^{1*}$ ~~~~Shaoshuai Shi$^{4}$ ~~~~Guangming Lu$^{1}$ ~~~~Daojing He$^{1}$ \\
~~~~Wenjie Pei$^{1,2}$\textsuperscript{\Letter} ~~~~Li Jiang$^{3}$\textsuperscript{\Letter}\\
{{$^1$}Harbin Institute of Technology, Shenzhen} ~~{{$^2$}Pengcheng Laboratory}\\
{{$^3$}The Chinese University of Hong Kong, Shenzhen} ~~{{$^4$}Voyager Research, Didi Chuxing}\\
{\tt\small wenjiecoder@outlook.com~~jiangli@cuhk.edu.cn }\\}

\begin{document}

\maketitle


\blfootnote{\textsuperscript{\dag}Research Assistant at CUHKSZ.}
\blfootnote{\textsuperscript{*}Equal contribution to this work. ~\textsuperscript{\Letter} Corresponding author.}

\input{sec/0_abstract}    
\input{sec/1_intro}
\input{sec/2_related_works}
\input{sec/3_methods}
\input{sec/4_experiments}
\input{sec/5_conclusions}
{
    \small
    \bibliographystyle{ieeenat_fullname}
    \bibliography{main}
}

\input{X_suppl}


\end{document}

%% file: sec/0_abstract.tex
\begin{abstract}
Recent advances in point cloud perception have demonstrated remarkable progress in scene understanding through vision-language alignment leveraging large language models (LLMs). However, existing methods may still encounter challenges in handling complex instructions that require accurate spatial reasoning, even if the 3D point cloud data provides detailed spatial cues such as size and position for identifying the targets. To tackle this issue, we propose Relevant Reasoning Segmentation (R$^2$S), a reasoning-based segmentation framework. The framework emulates human cognitive processes by decomposing spatial reasoning into two sequential stages: first identifying relevant elements, then processing instructions guided by their associated visual priors. Furthermore, acknowledging the inadequacy of existing datasets in complex reasoning tasks, we introduce 3D ReasonSeg, a reasoning-based segmentation dataset comprising 25,185 training samples and 3,966 validation samples with precise annotations. Both quantitative and qualitative experiments demonstrate that the R$^2$S and 3D ReasonSeg effectively endow 3D point cloud perception with stronger spatial reasoning capabilities, and we hope that they can serve as a new baseline and benchmark for future work.
\end{abstract}

%% file: sec/1_intro.tex
\begin{figure*}[h]
\centering
\includegraphics[width=0.95 \linewidth]{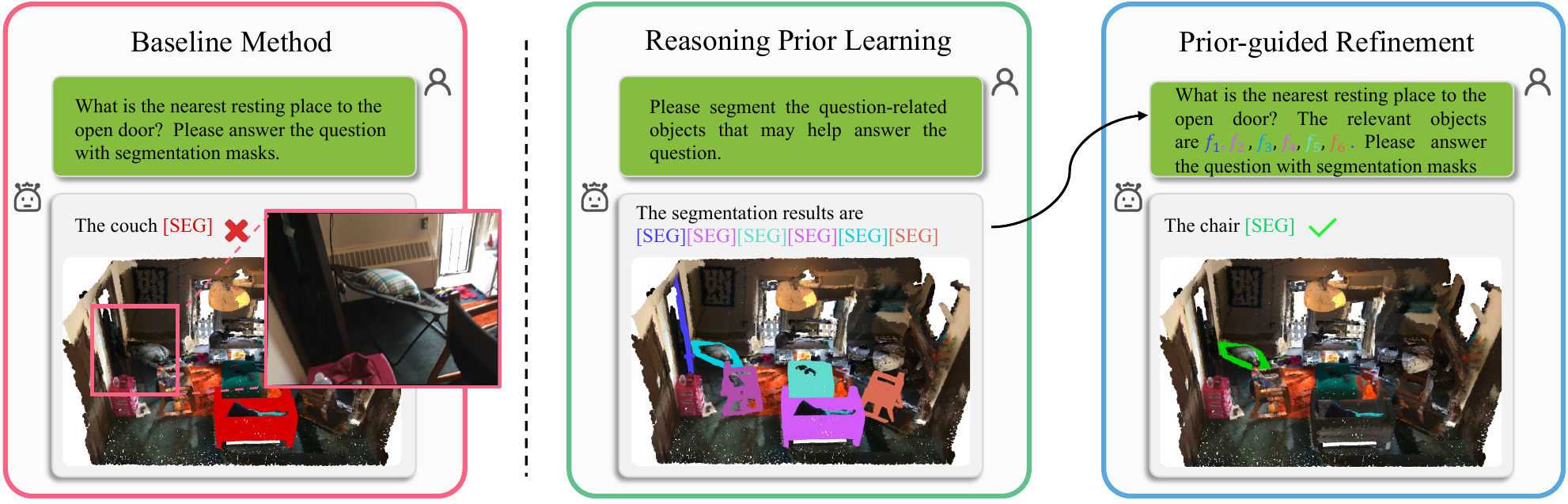}
\caption{In contrast to the baseline model in Sec. 3.1, we introduce Relevant Reasoning Segmentation to explicitly guide the model’s reasoning by emphasizing target-relevant objects. In this instance, the object identified as the nearest resting place to the open door should be the chair marked by the red rectangle. The first to third columns in the image represent the prediction from the base method, the outcomes of reasoning prior learning, and the final prediction following prior guided refinement.}
\label{fig:teaser}
\end{figure*}

\section{Introduction}
\label{sec:intro}

In recent years, multi-modal large language models (MLLMs)~\citep{alayrac2022flamingo,li2023blip,ye2023mplug,zhu2023minigpt,yang2023mm,wu2023visual} have managed to connect language and vision, bringing in a new era of artificial intelligence. These models have shown unprecedented abilities, enabling various fields to understand and create textual and visual content, such as human-computer interaction, multimedia understanding, embodied intelligence, and more. 

Building upon these foundations, recent developments in MLLMs have led to the emergence of 3D MLLMs~\citep{chen2024ll3da,hong20233d,wang2023chat,huang2023chat,xu2023pointllm}, which demonstrate remarkable capabilities in 3D point cloud perception through enhanced vision-language alignment. Despite these achievements, current 3D MLLM-based approaches exhibit limitations in processing instructions that require sophisticated spatial reasoning. As shown in \cref{fig:teaser}, the baseline model fails to accurately localize the target object. This observation raises an important question: while 3D point clouds inherently contain richer spatial information (\eg, precise measurements of size and position) compared to 2D images, why do models still struggle to effectively leverage these geometric properties? We posit that the development of robust spatial reasoning capabilities requires two critical components: an explicit reasoning mechanism and a dataset that encompasses complex reasoning scenarios.

In this work, we introduce the Relevant Reasoning Segmentation (R$^2$S), a reasoning-based segmentation framework that enhances the spatial reasoning capabilities of 3D MLLMs through two key steps: 1) Reasoning Prior Learning and 2) Prior-guided Refinement. In the first step, the model is trained to generate spatial priors by identifying target-relevant objects, providing essential contextual information. In the second step, the model leverages these reasoning priors to refine hidden representations, facilitating precise target localization. In other words, the two steps emulate human reasoning behavior, \ie, initially identifying relevant elements broadly and then scrutinizing them closely to pinpoint the target.

Analysis of existing datasets reveals that instructions requiring reasoning are often oversimplified and ambiguous, as illustrated in \cref{fig:ambiguous_case}. This limitation constrains the training and evaluation processes for developing reasoning capabilities in 3D MLLMs. To address this limitation, we present 3D ReasonSeg, a reasoning-based segmentation dataset that focuses on object functionality, visual attributes, and spatial relationships. The dataset comprises 25,185 training samples and 3,966 validation samples, with quality assurance maintained through manual verification processes. This dataset facilitates enhancement and comprehensive assessment of spatial reasoning capabilities.

Both our proposed R$^2$S framework and the 3D ReasonSeg dataset contribute to substantial performance improvements on established benchmarks, demonstrating their complementary effects in enhancing spatial reasoning capabilities. In summary, our main contributions are:

\begin{itemize}[leftmargin=0.7cm]
\item We propose the Relevant Segmentation (R$^2$S), a reasoning-based segmentation framework that enhances spatial reasoning capabilities of 3D MLLMs by emulating human reasoning processes through two key components: Reasoning Prior Learning and Prior-guided Refinement.

\item We propose 3D ReasonSeg, a reasoning-based segmentation dataset that focuses on object functionality, visual attributes, and spatial relationships, designed to enhance and comprehensively assess the spatial reasoning capabilities in 3D MLLMs.

\item Extensive experimental results demonstrate that our R$^2$S framework, combined with the 3D ReasonSeg dataset, achieves significant improvements in spatial reasoning across multiple benchmarks.
\end{itemize}

%% file: sec/2_related_works.tex
\section{Related Work}
\subsection{3D Scene Understanding}
3D scene understanding requires models to detect objects and comprehend their characteristics in 3D space~\citep{chen2023trajectoryformer,chen2022mppnet,cai2021semantic,kolodiazhnyi2024oneformer3d,shi2019pointrcnn,misra2021end}. Among these tasks, 3D segmentation presents unique challenges as it demands point-wise classification within point clouds. Recent studies demonstrate that language integration enhances model performance, as evidenced in 3D Question Answering and 3D Referring tasks~\citep{azuma2022scanqa,ma2022sqa3d,chen2020scanrefer,zhang2023multi3drefer,huang2022multi}. These language-guided tasks, which require models to respond to instructions or identify objects based on descriptions, underscore the importance of language comprehension and spatial reasoning capabilities.

The integration of language has supported models in transitioning from closed-set to open-set scenarios, such as in 3D Open Vocabulary Segmentation~\citep{peng2023openscene,das2024mta,chen2020scanrefer,ding2023pla,takmaz2023openmask3d}. This step is critical as it allows the model to break free from being constrained by the finite set of categories in the training dataset. While most task-oriented studies focus on specific objectives, recent advancements in 3D multi-modal large language models have enabled addressing multiple 3D downstream tasks using these models~\citep{chen2024ll3da,huang2023embodied}. Furthermore, the assistance of 3D multi-modal large language models has empowered models to tackle more intricate object localization tasks~\cite{zhu2024empowering,chen2024grounded}. 

These prior studies have not adequately addressed the complex spatial reasoning capabilities required for comprehensive understanding of 3D scenes. In this work, we introduce Relevant Reasoning Segmentation and 3D ReasonSeg to address complex spatial reasoning challenges.

\subsection{3D Multi-modal Large Language Model}
Significant progress has been made in the realm of 3D multi-modal large language models (3D MLLMs). Initial endeavors have predominantly concentrated on point clouds at the object level~\citep{han2023imagebind,guo2023point,xu2023pointllm,liu2024uni3d}, posing challenges in tackling complex tasks at the scene level. Recently, researchers have shifted attention to understanding entire scene point clouds~\citep{chen2024ll3da,wang2023chat,huang2023chat,hong20233d}, significantly advancing 3D scene understanding.

Drawing from these advances, we present an efficient framework that handles various scene understanding tasks, with particular strength in segmentation. While existing 3D MLLMs can process basic spatial relationships, they struggle with complex spatial reasoning tasks. Our model, however, demonstrates superior performance in handling sophisticated spatial reasoning challenges.

%% file: sec/3_methods.tex
\begin{figure*}[t]
\centering
\includegraphics[width=0.95 \linewidth]{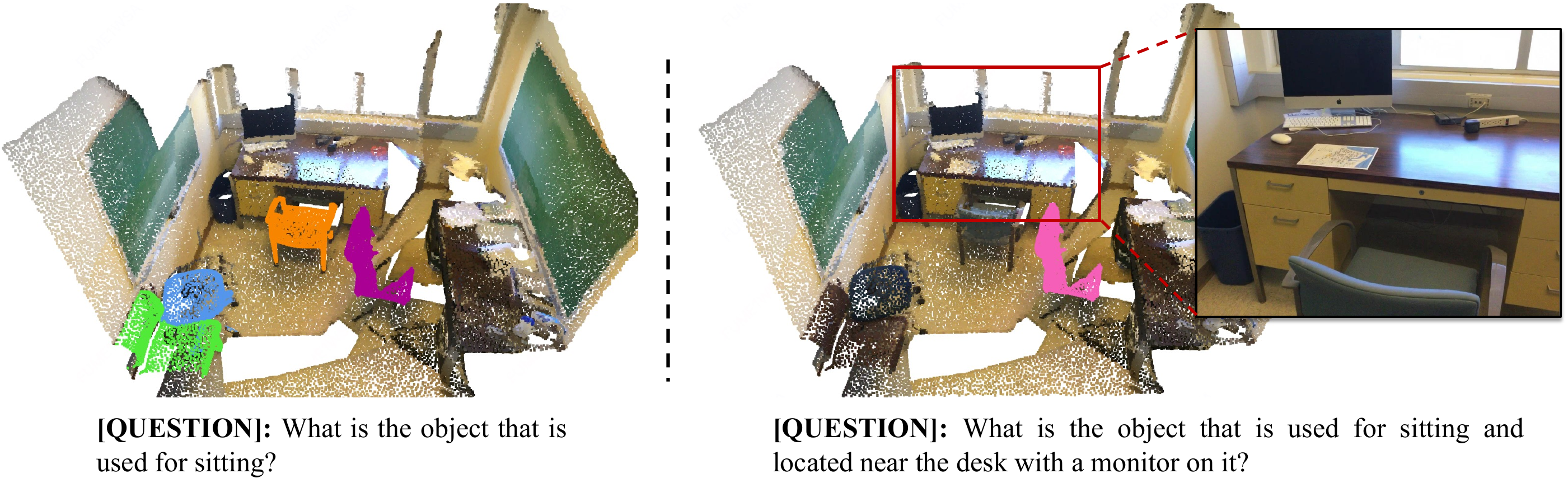}
\caption{The image on the left depicts the LLM's response to the question ``\textit{What is the object that is used for sitting?}'' Conversely, the right side of the image illustrates the LLM's response to the question ``\textit{What is the object that is used for sitting and located \mypara{near the desk with a monitor on it}}'' It is noteworthy that the LLM provides an incorrect answer in the latter case. The accurate response should identify the chair within the red rectangle.}
\label{fig:need_reasoning_case}
\end{figure*}

\section{Method}
Efforts such as those outlined in~\citep{chen2024ll3da, hong20233d, huang2023chat} were dedicated to leveraging 3D MLLMs for a deeper comprehension of 3D point clouds. Nevertheless, it is evident that current methods encounter challenges in effectively managing intricate spatial relationships. Specifically, as illustrated in \cref{fig:need_reasoning_case}, while the model can accurately identify "What is the object that is used for sitting?", it struggles with queries such as "What is the object that is used for sitting and located near the desk with a monitor on it?"

To tackle this issue, in this work, we instruct the model to generate reasoning priors by paying attention to objects relevant to the target. Subsequently, the model examines the scene and these relevant objects more closely to locate the final target. We refer to this process as Relevant Reasoning Segmentation (R$^2$S).

The remainder of this section is organized as follows: \cref{sec:model_architecture} introduces our baseline model, which demonstrates strong dense perception capabilities on 3D point cloud data. \cref{sec:spatial_relation_reasoning} elaborates on our proposed Relevant Reasoning Segmentation framework. \cref{sec:data_preparation} details the construction of our 3D ReasonSeg dataset, and \cref{sec:model_training} describes the comprehensive training pipeline.

\begin{figure*}[t]
\centering
\resizebox{0.95 \linewidth}{!}{\includegraphics{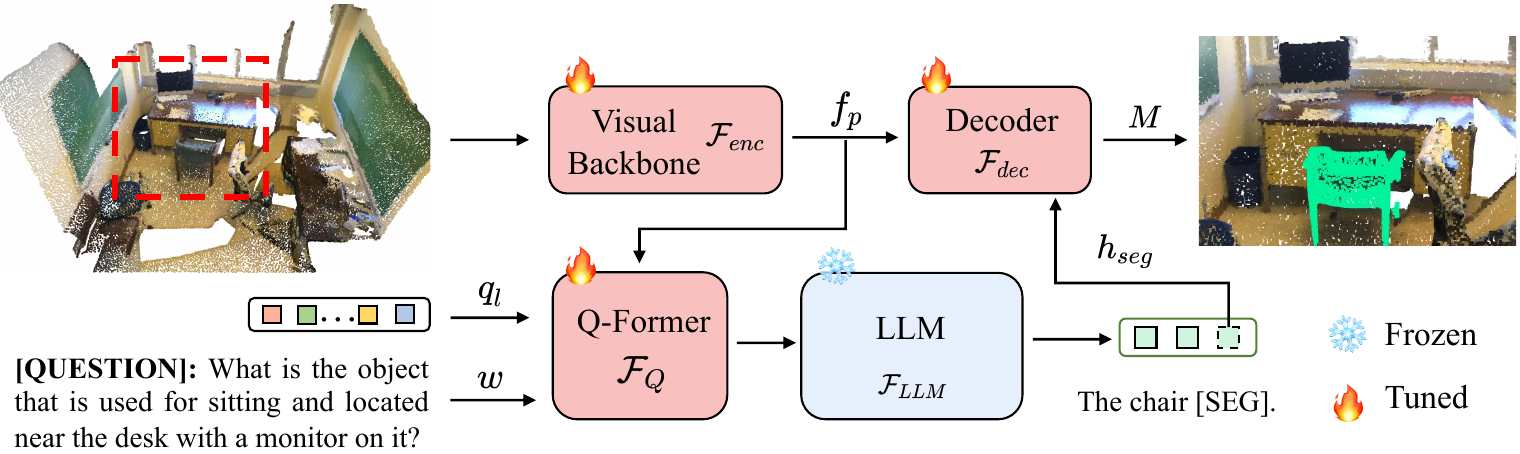}}
\caption{The baseline model consists of a visual backbone for point feature encoding, a Q-Former that connects visual-language representations, a LLM for instruction processing, and a mask decoder that converts \texttt{[SEG]} tokens into segmentation masks.}
\label{fig:architecture}
\end{figure*}

\subsection{Baseline Model}
\label{sec:model_architecture}
Our baseline architecture consists of a multi-modal large language model (MLLM) and a Segmentation Head. Following~\cite{lai2024lisa}, we extend the MLLM's vocabulary with a special token \texttt{[SEG]} that instructs the Segmentation Head to generate corresponding segmentation mask. This design enables the MLLM to effectively perform fine-grained segmentation tasks while maintaining its core language understanding capabilities.

\textbf{Multi-modal Large Language Model.} There are three key components in the adopted MLLM for feature extraction: a visual encoder, a Q-Former, and a large language model (LLM). Initially, the point cloud $X_{p}$ is processed through the visual encoder to extract dense features. 
To optimize computational efficiency, we employ super-point~\cite{landrieu2018large} within the visual backbone, aggregating the dense visual features into super-point features $f_{p}$ to reduce the following computational burden. Subsequently, the Q-Former~\cite{li2023blip} compresses the scene's information into several latent queries $\vec{q}_{l}$ 
based on the super-point features $\vec{f}_{p}$ and text embedding $\vec{w}$ of the instruction. 
After that, the latent queries $\vec{\hat{q}}_l$ will serve as the visual input, along with the text embedding $\vec{w}$, to be processed by the large language model $\mathcal{F}_{LLM}$. This process is formulated as:

\begin{equation}
\label{eq:base_multimodalLLM_first}
    \vec{f}_{p} = \mathcal{F}_{enc}(\vec{X}_{p}),
    \hspace{0.3 cm}
    \vec{\hat{q}}_l = \mathcal{F}_{Q}(\vec{q}_l, \vec{w}, \vec{f}_{p}).
\end{equation}

\begin{equation}
\label{eq:base_multimodalLLM_second}
    \vec{y}_{txt} = \mathcal{F}_{LLM}(\vec{\hat{q}}_l, \vec{w}).
\end{equation}

Inside the Q-Former $\mathcal{F}_{Q}$, $\vec{q}_l$ is concatenated with $\vec{w}$ to be the query in both the self-attention and cross-attention layers of $\mathcal{F}_{Q}$, while $\vec{f}_p$ serves as the key and value in the cross-attention layer:
\begin{gather}
\label{eq:qformer_forward_base}
\vec{q}_l, \vec{w} = \texttt{SelfAttn}(\texttt{Concat}(\vec{q}_l, \vec{w})), \\
\vec{q}_l, \vec{w} = \texttt{CrossAttn}(\texttt{Concat}(\vec{q}_l, \vec{w}), \vec{f}_p, \vec{f}_p).
\end{gather}

\textbf{Segmentation Head.} 
If the output of LLM, \ie, $\vec{y}_{txt}$, contains \texttt{[SEG]} tokens, it means that the current query requires segmentation predictions.
Then, the corresponding hidden features of the \texttt{[SEG]} token, \ie, $\vec{h}_{seg}$, are processed by the segmentation head $\mathcal{F}_{dec}$ that follows the decoder structure of~\cite{segmentanything_sam} to yield the requisite segmentation mask $M$, based on the reasoning between $\vec{h}_{seg}$ and $\vec{f}_p$. The process can be formulated as follows.

\begin{equation}
    \vec{M} = \mathcal{F}_{dec}(\vec{h}_{seg}, \vec{f}_p).
\end{equation}

The aforementioned framework seamlessly incorporates dense visual perception abilities into the output of the language model, allowing for the fusion of linguistic and visual information processing.

\subsection{Relevant Reasoning Segmentation}
\label{sec:spatial_relation_reasoning}

The baseline model can decently handle point cloud perception tasks. However, in our observations, the direct instruction tuning performed in~\cite{lai2024lisa} did not consider the spatial relationships among objects in the scene, hindering complex spatial reasoning, as illustrated in \cref{fig:qualitative_results}. Hence, to address this issue, we suggest that the model initially identifies the objects related to the target and their spatial relationships, serving as the reasoning prior to guide the target localization. This process is thus termed as Relevant Reasoning Segmentation (R$^2$S), and it contains two steps as follows.

\textbf{Step 1: Reasoning Prior Learning}. While the model may encounter challenges in understanding intricate inter-object relationships, it can recognize objects relevant to the target. As shown in \cref{fig:need_reasoning_case}, despite the model's limitation in accurately segmenting the chair beside the desk supporting a monitor, it demonstrates the ability to detect desks and chairs. Given this capability, we may instruct the model to initially segment approximate question-related objects, thus establishing the visual reasoning prior as the hint to help identify the final target. 
 
Following the baseline model's process, we first use an instruction structured as: ``\texttt{Given the 3D scene, \texttt{[QUESTION]}. Please segment the question-related objects that may help answer the question.}"~\footnote{More templates are shown in the supplementary file}. The corresponding instruction embedding $\vec{w}$ will be processed by \cref{eq:base_multimodalLLM_first} and \cref{eq:base_multimodalLLM_second} to obtain the text output $\vec{y}_{txt}$ from which we can obtain the \texttt{[SEG]} tokens representing these target-relevant objects. 
Subsequently, the segmentation head decodes the hidden features $\vec{h}_{seg}$ of \texttt{[SEG]} tokens into masks $\vec{M}_{r}$ for these target-relevant objects. 

Next, we adopt mask-pooling to aggregate the super-point features $\vec{f}_{p}$ with masks $\vec{M}_{r}$, yielding the representations $\vec{f}_{r}$ for the target-relevant objects. The features $\vec{f}_{r}$ serve as the reasoning priors in the subsequent step. 
The above process can be formally expressed as:

\begin{equation}
    \vec{M}_{r} = \mathcal{F}_{dec}(\vec{f}_{p}, \vec{h}_{seg}),
\end{equation}
\begin{equation}
\label{eq:prior_generation}
    \vec{f}_{r} = \vec{f}_{p} \times \vec{M}_{r}
\end{equation}

\textbf{Step 2: Prior-guided Refinement.} 
In the absence of explicit instructions, the latent queries $\vec{\hat{q}}_{l}$ derived from \cref{eq:base_multimodalLLM_first} may exhibit a bias towards irrelevant areas, potentially impacting the reasoning process in \cref{eq:base_multimodalLLM_second}. Besides, the mask-pooling operation in \cref{eq:prior_generation} may cause the loss of essential spatial details.
Therefore, we propose the prior-guided refinement which commences by feeding the initial latent queries $\vec{q}_l$, text outputs $y_{txt}$ and target-relevant features $\vec{f}_r$ from step 1 into the Q-Former $\mathcal{F}_{Q}$ to yield the refined query $ \vec{q}'_l$ and target-relevant features $\vec{f}'_r$. 

\begin{figure*}[t]
\centering
\resizebox{0.98 \linewidth}{!}{\includegraphics{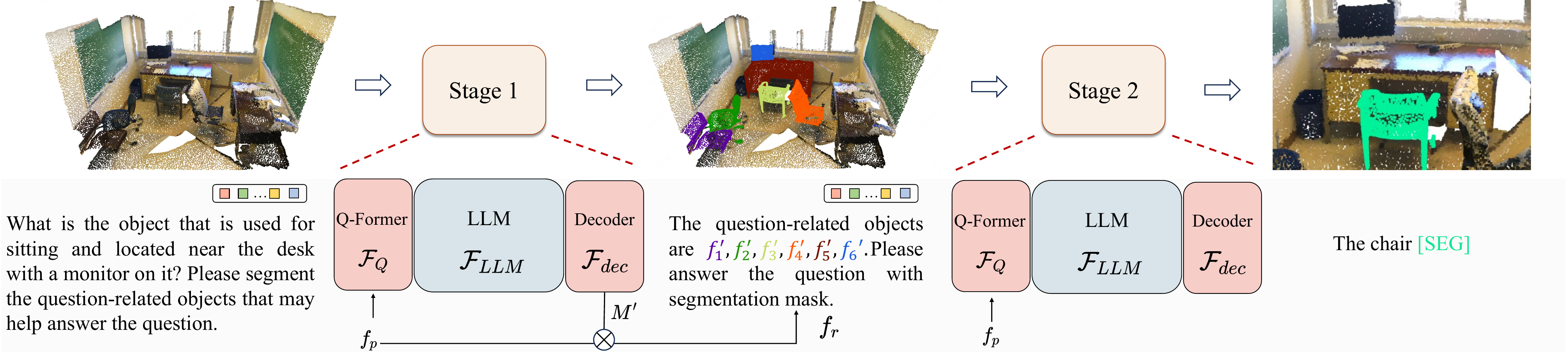}}
\caption{The proposed Relevant Reasoning Segmentation framework. Stage 1 segments question-related objects as priors ($f^{'}_{1},..,f^{'}_{6}$). Stage 2 refines hidden representations of instructions based on these priors, enabling the LLM to generate more accurate responses.}
\label{fig:spatial_relation_reasoning}
\end{figure*}

To make use of the obtained reasoning priors, we employ a revised instruction like: ``\texttt{Given the 3D scene, the question is [QUESTION]. Please answer the question and output segmentation mask. The question-related objects are \{$\vec{f}_{r}^1$, $\vec{f}_{r}^2$, ..., $\vec{f}_{r}^n$\}, and you may need to pay attention to them}''~\footnote{More templates are shown in the supplementary file.} where the target-relevant features $\vec{f}_{r}$ of $n$ potentially relevant objects have been inserted to form the new instruction whose text embedding is therefore denoted as $\vec{w}'\oplus\vec{f}_{r}$. To this end, the prior-guided refinement regarding the latent query and the target-relevant features can be formulated as:
\begin{equation}
\label{eq:prior_guided_refine}
    \vec{q}'_l, \vec{f}'_r = \mathcal{F}_{Q}(\vec{q}_{l}, \vec{w}'\oplus\vec{f}_{r}, \vec{f}_{p})
\end{equation}
where $\vec{w}'$ and $\vec{f}_{p}$ are the text embeddings and super-point features, respectively. Unlike \cref{eq:qformer_forward_base}, during the prior-guided refinement, $\vec{q}_l$ is concatenated with the instruction embedding $\vec{w}'\oplus\vec{f}_{r}$ to serve as the queries for both self-attention and cross-attention layers in the Q-Former $\mathcal{F}_{Q}$. Meanwhile, $\vec{f}_p$ continues to function as the key and value in the cross-attention layer.
In \cref{eq:prior_guided_refine}, the enhanced latent query $\vec{q}'_l$ has been enriched with target-related clues from $\vec{f}_{r}$. Simultaneously, utilizing the point-cloud features $\vec{f}_{p}$, the target-relevant features $\vec{f}_{r}$ are refined with essential spatial intricacies to produce $\vec{f}'_r$.
Next, the refined query $ \vec{q}'_l$ and target-relevant features $\vec{f}'_r$ are adopted for generating the new text output $\vec{y}'_{txt}$ with the LLM $\mathcal{F}_{LLM}$:
\begin{equation}
    \vec{y}'_{txt} = \mathcal{F}_{LLM}(\vec{q}'_{l}, \vec{w}'\oplus\vec{f}'_r),
\end{equation}
where the instruction embedding $\vec{w}'\oplus\vec{f}_{r}$ originally used in \cref{eq:prior_guided_refine} has been accordingly updated to $\vec{w}'\oplus\vec{f}'_r$.
The hidden features $\vec{h}'_{seg}$ of the \texttt{[SEG]} tokens are extracted from the new text output $\vec{y}'_{txt}$ and processed by the decoder $\mathcal{F}_{dec}$ to generate the refined final mask prediction $\vec{M}'$:
\begin{equation}
    \vec{M}' = \mathcal{F}_{dec}(\vec{h}'_{seg}, \vec{f}_p)
\end{equation}

\begin{figure}[h]
\centering
\includegraphics[width=0.85 \linewidth]{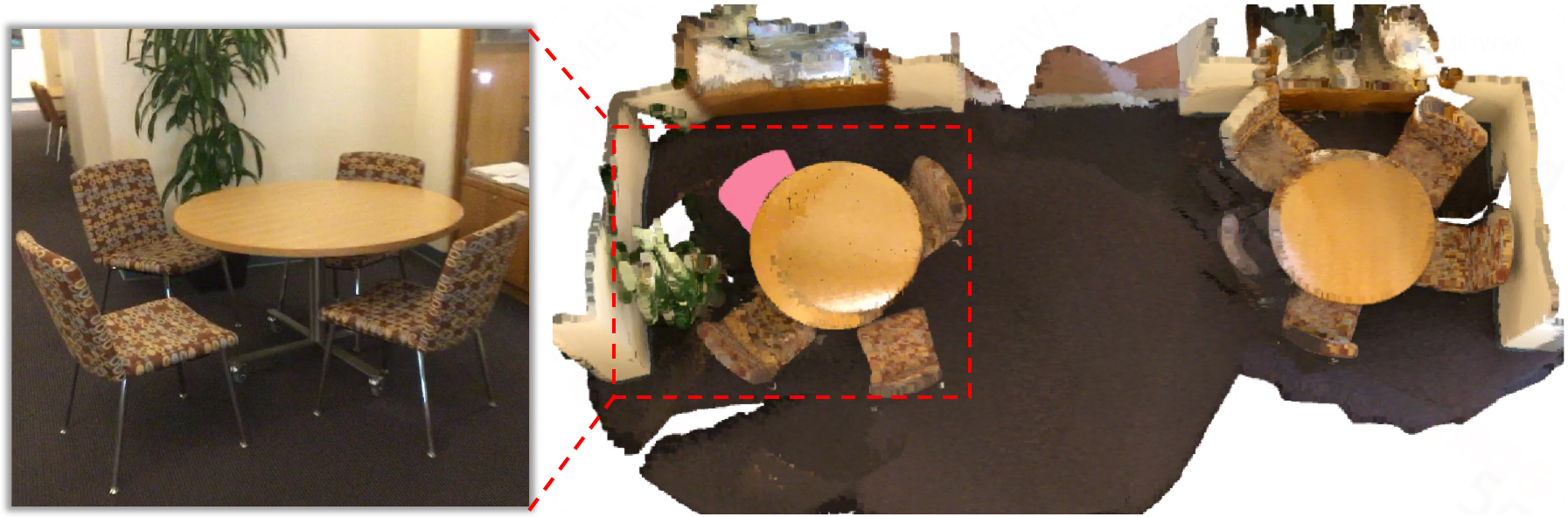}
\caption{This illustrates a data sample from ScanRefer, where the target object is described as "A brown chair placed in front of a plant." The spatial relationship in this description shows ambiguity, as there are multiple chairs in front of the plant.}
\label{fig:ambiguous_case}
\end{figure}

\subsection{3D ReasonSeg}
\label{sec:data_preparation}

Analysis of existing datasets, as demonstrated in \cref{fig:ambiguous_case}, reveals that spatial relations such as "in front of" often lack complexity and exhibit potential ambiguity. Using such datasets for training and evaluation may inadequately address the spatial reasoning requirements of 3D MLLMs. To facilitate enhancement and assessment of spatial reasoning capabilities in 3D MLLMs, we present 3D ReasonSeg, a dataset specifically designed for reasoning tasks in 3D point cloud environments. Empirical results in \cref{tab:main_results} indicate that models trained on 3D ReasonSeg achieve substantial performance improvements over baseline implementations.

Specifically, the proposed 3D ReasonSeg requires the model to identify the target object based on a question that does not explicitly mention it. The model needs to understand the object functions, visual attributes, and spatial relationships of the objects described in the query and reason to segment the target object. To implement this task, we use a template structured as: ``\texttt{\textbf{USER:} Given the 3D scene, please answer the question and output segmentation mask: [QUESTION]. \textbf{ASSISTANT:} [ANSWER]}".

We construct the 3D ReasonSeg dataset by integrating SceneVerse's~\cite{jia2024sceneverse} detailed object descriptions with LLama 3.1~\cite{dubey2024llama3herdmodels}. Based on objects' 3D spatial information (positions and sizes) from point clouds and the object descriptions from SceneVerse, LLama 3.1 generates question-reasoning-answer pairs for reasoning tasks.

After collecting numerous question-reasoning-answer pairs, we apply rule-based filtering (e.g., calculating distances between objects and removing outliers) to eliminate instances with unclear or incorrect spatial relationships. The resulting 3D ReasonSeg dataset contains 29,151 high-quality data samples. Examples are shown in \cref{fig:dataset_sample}, with details provided in the supplementary material.

We conduct a statistical analysis of ScanQA, ScanRefer, and 3D ReasonSeg datasets, examining their training (\#Train) and validation (\#Val) set sizes, average word counts per sample (Avg. Words), and average number of relevant objects (Avg. Objects). The detailed statistics are summarized in \cref{tab:statistic_of_reasonseg}, showing that our proposed 3D ReasonSeg dataset contains more complex object relationships.

\begin{table}[h]
\centering
\resizebox{0.95\linewidth}{!}{
\begin{tabular}{c|cccc}
\toprule 
\multicolumn{1}{c|}{Dataset} & \multicolumn{1}{c}{\#Train} & \multicolumn{1}{c}{\#Val} & {Avg. Words} & \multicolumn{1}{c}{Avg. Objects} \\ \hline
\multicolumn{1}{c|}{ScanQA} & \multicolumn{1}{c}{26,563} & \multicolumn{1}{c}{4,675} & \multicolumn{1}{c}{8.8} & \multicolumn{1}{c}{1.5} \\
\multicolumn{1}{c|}{ScanRefer} & \multicolumn{1}{c}{36,665} & \multicolumn{1}{c}{9,508} & \multicolumn{1}{c}{17.8} & \multicolumn{1}{c}{1.8} \\
\multicolumn{1}{c|}{3D ReasonSeg} & \multicolumn{1}{c}{25,185} & \multicolumn{1}{c}{3,966} & \multicolumn{1}{c}{19.6} & \multicolumn{1}{c}{\bf 5.4} \\
\bottomrule
\end{tabular}
}
\caption{Comparative statistics across ScanQA, ScanRefer, and 3D ReasonSeg datasets.}
\label{tab:statistic_of_reasonseg}
\end{table}

\begin{figure}[h]
\centering
\resizebox{0.9 \linewidth}{!}{
\includegraphics[]{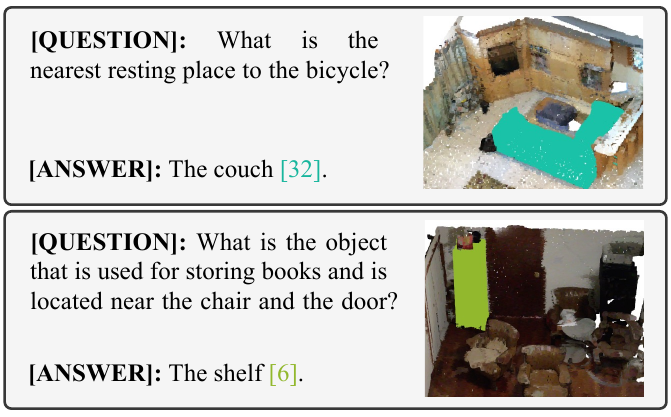}
}
\caption{Visualization of 3D ReasonSeg samples, where instance ID colors correspond to their respective object colors in the image.}
\label{fig:dataset_sample}
\end{figure}

\subsection{Model Training}
\label{sec:model_training}

\mypara{Training Datasets.} Following~\cite{chen2024ll3da}, our training data includes established 3D datasets such as ScanQA~\cite{azuma2022scanqa}, ScanRefer~\cite{chen2020scanrefer}, ScanNet200~\cite{dai2017scannet}, the ScanNet subset of 3D-LLM~\cite{hong20233d}, and SceneVerse~\cite{jia2024sceneverse}, along with our proposed 3D ReasonSeg dataset.

For Relevant Reasoning Segmentation outlined in \cref{sec:spatial_relation_reasoning}, we extend the ScanQA, Scanrefer, and 3D ReasonSeg to incorporate target-relevant object identification. This process leverages scene object annotations and question-answer pairs from the original datasets, with LLama 3.1~\cite{dubey2024llama3herdmodels} serving as the backbone for target-relevant object identification. Representative examples are presented in \cref{fig:target_relevant_objects_case}, with detailed prompts and implementation procedures in the supplementary material.

\mypara{Loss Function.} During training, the model is optimized using the cross-entropy loss $\mathcal{L}_{txt}$ for text generation, along with the binary cross-entropy (BCE) and DICE losses ($\mathcal{L}_{bce}$ and $\mathcal{L}_{dice}$) for segmentation. Following~\cite{kolodiazhnyi2024oneformer3d}, we assign equal weights of 1 to the BCE and DICE loss. Thus, the overall training objective $\mathcal{L}$ can be expressed as:
\begin{equation}
\label{training_objectives}
    \mathcal{L} =  \lambda_{txt}\mathcal{L}_{txt} + \mathcal{L}_{bce} + \mathcal{L}_{dice}
\end{equation}
where $\lambda_{txt}$ is the weight for text generation loss.

\mypara{Relevant Objects Augmentation.} Our relevant segmentation leverages mask predictions of relevant objects as priors to enhance latent query and text embedding. However, direct implementation leads to overfitting during training, as the ground truth relevant objects remain fixed during training but vary during inference, creating a train-test discrepancy. To address this issue, we implement random omission and addition of relevant objects during training, simulating diverse inference scenarios with incomplete priors.

\begin{figure}[h]
\centering
\resizebox{0.95 \linewidth}{!}{
\includegraphics[]{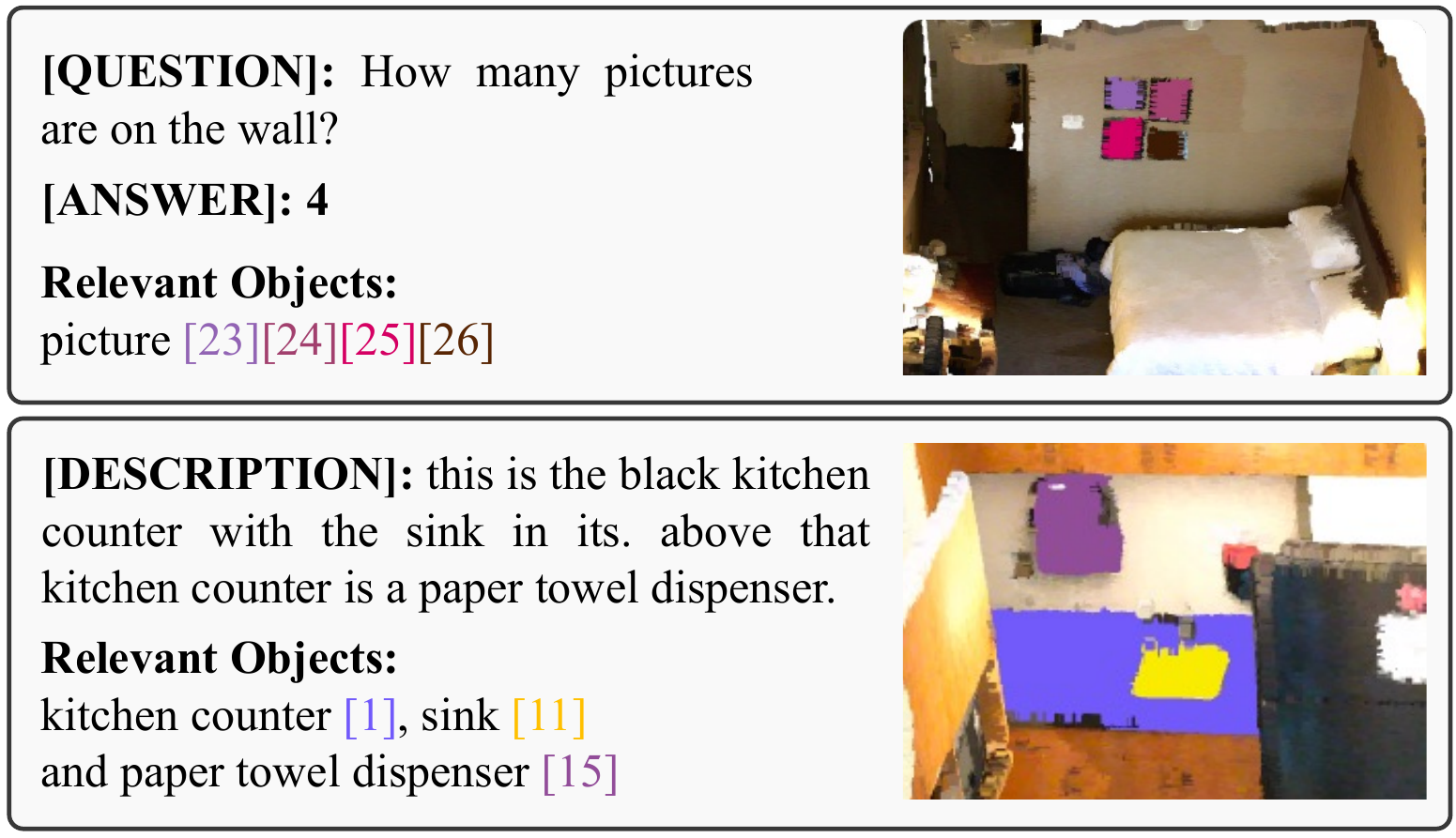}
}
\caption{Examples of target-relevant data. The first row shows an instance from ScanQA, and the second row is related to ScanRefer. The visualizations of target-relevant objects are displayed on the right side of each row.}
\label{fig:target_relevant_objects_case}
\end{figure}

%% file: sec/4_experiments.tex
\begin{table*}[h]
\centering
\resizebox{0.85 \linewidth}{!}{
\begin{tabular}{c|cc|cccc|c}
\toprule 
\multicolumn{1}{c|}{\multirow{2}{*}{Method}} & \multicolumn{2}{c|}{ScanRefer} & \multicolumn{4}{c|}{ScanQA} &\multicolumn{1}{c}{3D ReasonSeg}
\\
\cline{2-8}
\multicolumn{1}{l|}{} & \multicolumn{1}{c}{Acc@25} & \multicolumn{1}{c|}{Acc@50} & \multicolumn{1}{c}{B-4 $\uparrow$} & \multicolumn{1}{c}{C $\uparrow$} & \multicolumn{1}{c}{M $\uparrow$} & \multicolumn{1}{c}{R $\uparrow$} & \multicolumn{1}{|c}{gIoU}
\\ \hline
\multicolumn{1}{l|}{\color[HTML]{969696} ScanRefer ~\cite{chen2020scanrefer}} & \multicolumn{1}{c}{\color[HTML]{969696} 37.3} & \multicolumn{1}{c|}{\color[HTML]{969696} 24.3} & \multicolumn{1}{c}{\color[HTML]{969696} --} & \multicolumn{1}{c}{\color[HTML]{969696} --} & \multicolumn{1}{c}{\color[HTML]{969696} --} & \multicolumn{1}{c|}{\color[HTML]{969696} --} & \multicolumn{1}{c}{\color[HTML]{969696} --}
\\
\multicolumn{1}{l|}{\color[HTML]{969696}  MVT ~\cite{huang2022multi}} & \multicolumn{1}{c}{\color[HTML]{969696} 40.8} & \multicolumn{1}{c|}{\color[HTML]{969696} 33.3} & \multicolumn{1}{c}{\color[HTML]{969696} --} & \multicolumn{1}{c}{\color[HTML]{969696} --} & \multicolumn{1}{c}{\color[HTML]{969696} --} & \multicolumn{1}{c|}{\color[HTML]{969696} --} & \multicolumn{1}{c}{\color[HTML]{969696} --}
\\
\multicolumn{1}{l|}{\color[HTML]{969696} 3DVG-Trans ~\cite{zhao20213dvg}} & \multicolumn{1}{c}{\color[HTML]{969696} 45.9} & \multicolumn{1}{c|}{\color[HTML]{969696} 34.5} & \multicolumn{1}{c}{\color[HTML]{969696} --} & \multicolumn{1}{c}{\color[HTML]{969696} --} & \multicolumn{1}{c}{\color[HTML]{969696} --} & \multicolumn{1}{c|}{\color[HTML]{969696} --} & \multicolumn{1}{c}{\color[HTML]{969696} --}
\\
\multicolumn{1}{l|}{\color[HTML]{969696} ViL3DRel ~\cite{chen2022language}} & \multicolumn{1}{c}{\color[HTML]{969696} 47.9} & \multicolumn{1}{c|}{\color[HTML]{969696} 37.7} & \multicolumn{1}{c}{\color[HTML]{969696} --} & \multicolumn{1}{c}{\color[HTML]{969696} --} & \multicolumn{1}{c}{\color[HTML]{969696} --} & \multicolumn{1}{c|}{\color[HTML]{969696} --} & \multicolumn{1}{c}{\color[HTML]{969696} --}
\\
\multicolumn{1}{l|}{\color[HTML]{969696}  M3DRef-CLIP ~\cite{zhang2023multi3drefer}} & \multicolumn{1}{c}{\color[HTML]{969696} 51.9} & \multicolumn{1}{c|}{\color[HTML]{969696} 44.7} & \multicolumn{1}{c}{\color[HTML]{969696} --} & \multicolumn{1}{c}{\color[HTML]{969696} --} & \multicolumn{1}{c}{\color[HTML]{969696} --} & \multicolumn{1}{c|}{\color[HTML]{969696} --} & \multicolumn{1}{c}{\color[HTML]{969696} --}
\\
\multicolumn{1}{l|}{\color[HTML]{969696}  ScanQA ~\cite{azuma2022scanqa}} & \multicolumn{1}{c}{\color[HTML]{969696} --} & \multicolumn{1}{c|}{\color[HTML]{969696} --} & \multicolumn{1}{c}{\color[HTML]{969696} 10.1} & \multicolumn{1}{c}{\color[HTML]{969696} 64.9} & \multicolumn{1}{c}{\color[HTML]{969696} 13.1} & \multicolumn{1}{c|}{\color[HTML]{969696} 33.3} & \multicolumn{1}{c}{\color[HTML]{969696} --}
\\
\multicolumn{1}{l|}{\color[HTML]{969696}  3D-VisTA ~\cite{zhu20233d}} & \multicolumn{1}{c}{\color[HTML]{969696} 50.6} & \multicolumn{1}{c|}{\color[HTML]{969696} 45.8} & \multicolumn{1}{c}{\color[HTML]{969696} 13.1} & \multicolumn{1}{c}{\color[HTML]{969696} 72.9} & \multicolumn{1}{c}{\color[HTML]{969696} --} & \multicolumn{1}{c|}{\color[HTML]{969696} --} & \multicolumn{1}{c}{\color[HTML]{969696} --}
\\ \hline
\multicolumn{1}{l|}{ LLM-Grounder ~\cite{yang2024llm}} & \multicolumn{1}{c}{17.1} & \multicolumn{1}{c|}{5.3} & \multicolumn{1}{c}{--} & \multicolumn{1}{c}{--} & \multicolumn{1}{c}{--} & \multicolumn{1}{c|}{--} & \multicolumn{1}{c}{--}
\\

\multicolumn{1}{l|}{ 3D-LLM ~\cite{hong20233d}} & \multicolumn{1}{c}{30.3} & \multicolumn{1}{c|}{--} & \multicolumn{1}{c}{12.0} & \multicolumn{1}{c}{69.4} & \multicolumn{1}{c}{14.5} & \multicolumn{1}{c|}{35.7} & \multicolumn{1}{c}{--}
\\
\multicolumn{1}{l|}{Chat-3D ~\cite{wang2023chat}} & \multicolumn{1}{c}{--} & \multicolumn{1}{c|}{--} & \multicolumn{1}{c}{6.4}  & \multicolumn{1}{c}{53.2} & \multicolumn{1}{c}{11.9} & \multicolumn{1}{c|}{28.5} & \multicolumn{1}{c}{--}
\\
\multicolumn{1}{l|}{Chat-3D v2 ~\cite{huang2023chat}} & \multicolumn{1}{c}{35.9} & \multicolumn{1}{c|}{30.4} & \multicolumn{1}{c}{7.3} & \multicolumn{1}{c}{77.1} & \multicolumn{1}{c}{16.1} & \multicolumn{1}{c|}{\bf 40.1} & \multicolumn{1}{c}{--}
\\
\multicolumn{1}{l|}{LL3DA ~\cite{chen2024ll3da}} & \multicolumn{1}{c}{--} & \multicolumn{1}{c|}{--} & \multicolumn{1}{c}{13.5} & \multicolumn{1}{c}{76.8} & \multicolumn{1}{c}{15.9} & \multicolumn{1}{c|}{37.3} & \multicolumn{1}{c}{--}
\\
\multicolumn{1}{l|}{ScanReason ~\cite{zhu2024empowering}} & \multicolumn{1}{c}{53.1} & \multicolumn{1}{c|}{41.1} & \multicolumn{1}{c}{--} & \multicolumn{1}{c}{--} & \multicolumn{1}{c}{--} & \multicolumn{1}{c|}{--} & \multicolumn{1}{c}{--}
\\
\multicolumn{1}{l|}{Grounded 3D-LLM ~\cite{chen2024grounded}} & \multicolumn{1}{c}{47.9} & \multicolumn{1}{c|}{44.1} & \multicolumn{1}{c}{13.4} & \multicolumn{1}{c}{72.7} & \multicolumn{1}{c}{--} & \multicolumn{1}{c|}{--} & \multicolumn{1}{c}{25.6}
\\
\hline
\multicolumn{1}{l|}{baseline} & \multicolumn{1}{c}{54.6} & \multicolumn{1}{c|}{38.7} & \multicolumn{1}{c}{12.8} & \multicolumn{1}{c}{70.5} & \multicolumn{1}{c}{13.8} & \multicolumn{1}{c|}{34.6} & \multicolumn{1}{c}{-}
\\
\multicolumn{1}{l|}{+ 3D ReasonSeg} & \multicolumn{1}{c}{58.9} & \multicolumn{1}{c|}{43.6} & \multicolumn{1}{c}{13.3} & \multicolumn{1}{c}{75.2} & \multicolumn{1}{c}{15.5} & \multicolumn{1}{c|}{36.7} & \multicolumn{1}{c}{29.2}
\\
\multicolumn{1}{l|}{+ $\text{R}^2\text{S}$} & \multicolumn{1}{c}{\bf 59.5} & \multicolumn{1}{c|}{\bf 48.7} & \multicolumn{1}{c}{\bf 13.9} & \multicolumn{1}{c}{\bf 77.3} & \multicolumn{1}{c}{\bf 16.4} & \multicolumn{1}{c|}{38.1} & \multicolumn{1}{c}{\bf 33.1}
\\
\bottomrule
\end{tabular}
}
\caption{Results among our methods and previous related works. Entries in gray denote models specialized for specific datasets.}
\label{tab:main_results}
\end{table*}

\section{Experiments}

\subsection{Experimental Setting}
\mypara{Implementation Details.} We adopt the encoder of a pre-trained Oneformer3D as visual backbone and the decoder of it as the mask decoder, and a pre-trained OPT 1.3B~\cite{zhang2022opt} as the language model. To preserve the learned knowledge of the pre-trained language model, we completely freeze the parameters of the language model while fully training the parameters of all other modules. Note that, we found it is necessary to also fine-tune the visual backbone. We adopt the AdamW~\cite{loshchilov2017decoupled} optimizer with a weight decay of 0.1 and a learning rate decaying from 1e{-}4 to 1e{-}6 with a cosine annealing scheduler. The text generation loss weight $\lambda_{txt}$ is set to 0.5. The number of learnable queries in Q-Former is set to 32. We randomly sample 300k points from each 3D scene as the 3D input. All the experiments are conducted on eight Nvidia H20 GPUs.

\mypara{Benchmark and Metrics.} In this study, we assess our methods on ScanQA~\cite{azuma2022scanqa}, ScanRefer~\cite{chen2020scanrefer}, and our proposed 3D ReasonSeg. ScanQA is a dataset for the 3D question answering task, evaluating visual and spatial comprehension of 3D environments. The evaluation metrics of ScanQA include BLEU-4, CIDEr, METEOR, and Rouge-L. ScanRefer evaluates the capability to localize 3D object using a natural language description. Prediction accuracy is measured by the percentage of predictions with Intersection over Union (IoU) above 0.25 and 0.5 compared to ground truth masks. For 3D ReasonSeg, the model is required to focus on object attributes and relationships to reason, answering questions and localizing target objects. Following~\cite{lai2024lisa}, the general Intersection over Union (gIoU) metric is used, defined as the average of all per-scene Intersection-over-Unions (IoUs). Detailed formulations of these tasks can be found in the supplementary material.

\subsection{Main Results} We evaluate our model's effectiveness through comprehensive comparisons with state-of-the-art (SOTA) methods, encompassing both task-specific specialist models and multi-task generalist models. For a fair comparison, we finetune Grounded 3D-LLM~\cite{chen2024grounded}, the previous SOTA model on ScanRefer, on our proposed 3D ReasonSeg dataset and assess its performance on the 3D ReasonSeg benchmark.

As shown in \cref{tab:main_results}, our baseline model is comparable to existing approaches on both ScanRefer and ScanQA. While our baseline model demonstrates robust performance, the incorporation of 3D ReasonSeg and R$^2$S yields consistent substantial improvements across all metrics, validating the effectiveness of our approach. Notably, R$^2$S significantly enhances the Acc@50 metric on ScanRefer and the gIoU metric on 3D ReasonSeg, tasks that require robust spatial reasoning, aligning with our expectations.

\begin{table}[h]
\centering
\resizebox{\linewidth}{!}{
\begin{tabular}{cccccccc}
\toprule 
\multicolumn{1}{c|}{\multirow{2}{*}{Method}} &\multicolumn{2}{c|}{ScanRefer} &\multicolumn{4}{c}{ScanQA} &\multicolumn{1}{|c}{3D ReasonSeg}
\\
\cline{2-8}
\multicolumn{1}{c|}{} & \multicolumn{1}{c}{Acc@25} & \multicolumn{1}{c|}{Acc@50} & \multicolumn{1}{c}{B-4 $\uparrow$} & \multicolumn{1}{c}{C $\uparrow$} & \multicolumn{1}{c}{M $\uparrow$} & \multicolumn{1}{c}{R $\uparrow$} & \multicolumn{1}{|c}{gIoU}
\\
\hline
\multicolumn{1}{l|}{wo PR} & \multicolumn{1}{c}{58.2} & \multicolumn{1}{c|}{41.3} & \multicolumn{1}{c}{7.7} & \multicolumn{1}{c}{66.1} & \multicolumn{1}{c}{13.4} & \multicolumn{1}{c|}{33.2} & \multicolumn{1}{c}{27.5}
\\
\multicolumn{1}{l|}{text-based} & \multicolumn{1}{c}{58.4} & \multicolumn{1}{c|}{43.5} & \multicolumn{1}{c}{13.7} & \multicolumn{1}{c}{75.5} & \multicolumn{1}{c}{15.6} & \multicolumn{1}{c|}{36.9} & \multicolumn{1}{c}{29.8}
\\
\multicolumn{1}{l|}{R$^2$S} & \multicolumn{1}{c}{\bf 59.5} & \multicolumn{1}{c|}{\bf 48.7} & \multicolumn{1}{c}{\bf 13.9} & \multicolumn{1}{c}{\bf 77.3} & \multicolumn{1}{c}{\bf 16.4} & \multicolumn{1}{c|}{\bf 38.1} & \multicolumn{1}{c}{\bf 33.1}
\\
\bottomrule
\end{tabular}
}
\caption{Ablation study on the Prior-guided Refinement.}
\vspace{-0.5em}
\label{tab:ablation_study_PR}
\end{table}

\begin{table}[h]
\centering
\resizebox{0.95 \linewidth}{!}{
\begin{tabular}{cccccccc}
\toprule 
\multicolumn{1}{c|}{number of} &\multicolumn{2}{c|}{ScanRefer} &\multicolumn{4}{c}{ScanQA} &\multicolumn{1}{|c}{3D ReasonSeg} \\ \cline{2-8}
\multicolumn{1}{c|}{latent queries} & \multicolumn{1}{c}{Acc@25} & \multicolumn{1}{c|}{Acc@50} & \multicolumn{1}{c}{B-4 $\uparrow$} & \multicolumn{1}{c}{C $\uparrow$} & \multicolumn{1}{c}{M $\uparrow$} & \multicolumn{1}{c}{R $\uparrow$} & \multicolumn{1}{|c}{gIoU} \\ \hline
\multicolumn{1}{c|}{16} & \multicolumn{1}{c}{57.2} & \multicolumn{1}{c|}{40.5} & \multicolumn{1}{c}{14.0} & \multicolumn{1}{c}{75.3} & \multicolumn{1}{c}{15.2} & \multicolumn{1}{c|}{36.8} & \multicolumn{1}{c}{28.9}
\\
\multicolumn{1}{c|}{32} & \multicolumn{1}{c}{\bf 58.9} & \multicolumn{1}{c|}{\bf 43.6} & \multicolumn{1}{c}{13.3} & \multicolumn{1}{c}{\bf 75.2} & \multicolumn{1}{c}{\bf 15.5} & \multicolumn{1}{c|}{\bf 36.7} & \multicolumn{1}{c}{\bf 29.2}
\\
\multicolumn{1}{c|}{48} & \multicolumn{1}{c}{58.7} & \multicolumn{1}{c|}{41.6} & \multicolumn{1}{c}{\bf 14.5} & \multicolumn{1}{c}{74.3} & \multicolumn{1}{c}{15.1} & \multicolumn{1}{c|}{36.1} & \multicolumn{1}{c}{28.4}
\\
\bottomrule
\end{tabular}
}
\caption{Ablation study on the number of latent queries. The results are based on our base method.}
\vspace{-0.5em}
\label{tab:ablation_study_n_latent_queries}
\end{table}

\begin{table}[!h]
\centering
\resizebox{\linewidth}{!}{
\begin{tabular}{c|c|cc|cccc|c}
\toprule 
\multirow{2}{*}{$\lambda_{txt}$} &\multicolumn{2}{c|}{ScanRefer} &\multicolumn{4}{c}{ScanQA} &\multicolumn{1}{|c}{3D ReasonSeg}
\\
\cline{2-8}
\multicolumn{1}{c|}{} & \multicolumn{1}{c}{Acc@25} & \multicolumn{1}{c|}{Acc@50} & \multicolumn{1}{c}{B-4 $\uparrow$} & \multicolumn{1}{c}{C $\uparrow$} & \multicolumn{1}{c}{M $\uparrow$} & \multicolumn{1}{c}{R $\uparrow$} & \multicolumn{1}{|c}{gIoU}
\\
\hline
\multicolumn{1}{c|}{0.1} & \multicolumn{1}{c}{61.3} & \multicolumn{1}{c|}{43.9} & \multicolumn{1}{c}{12.3} & \multicolumn{1}{c}{68.2} & \multicolumn{1}{c}{14.3} & \multicolumn{1}{c|}{34.0} & \multicolumn{1}{c}{28.9}
\\
\multicolumn{1}{c|}{0.5} & \multicolumn{1}{c}{\bf 58.9} & \multicolumn{1}{c|}{\bf 43.6} & \multicolumn{1}{c}{\bf 13.3} & \multicolumn{1}{c}{\bf 75.2} & \multicolumn{1}{c}{\bf 15.5} & \multicolumn{1}{c|}{\bf 36.7} & \multicolumn{1}{c}{\bf 29.2}
\\
\multicolumn{1}{c|}{1.0} & \multicolumn{1}{c}{58.7} & \multicolumn{1}{c|}{41.4} & \multicolumn{1}{c}{13.2} & \multicolumn{1}{c}{72.6} & \multicolumn{1}{c}{15.2} & \multicolumn{1}{c|}{35.6} & \multicolumn{1}{c}{28.6}
\\
\bottomrule
\end{tabular}
}
\caption{Ablation study on the loss weight configurations, where $\lambda_{txt}$ means the weight of $\mathcal{L}_{txt}$.}
\label{tab:ablation_loss_weigths}
\end{table}

\begin{figure*}[h]
\centering
\resizebox{0.95 \linewidth}{!}{
\includegraphics[]{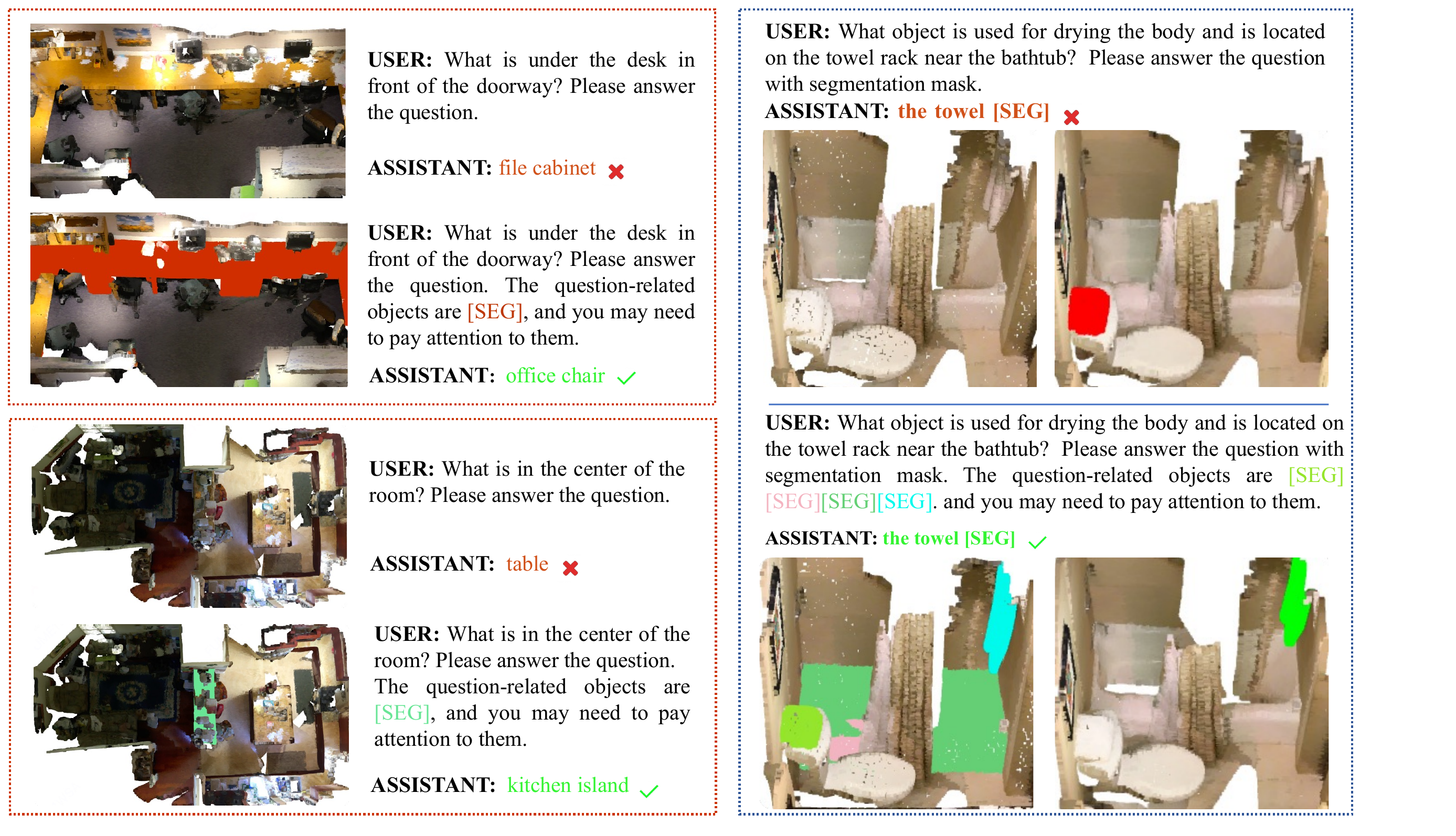}
}
\caption{Visual qualitative results on ScanQA and 3D ReasonSeg.}
\label{fig:qualitative_results}
\end{figure*}

\mypara{Effectiveness of Prior-guided Refinement.} To verify the necessity of Prior-guided Refinement (PR), we investigate the approach without PR (wo PR) and directly representing target-relevant objects as textual information (text-based). The results are shown in \cref{tab:ablation_study_PR}.

When Prior-guided Refinement is removed, the model simultaneously predict relevant objects and process instructions. Results indicate that the approach without Prior-guided Refinement not only fails to improve performance but potentially degrades it. This degradation likely stems from the increased computational complexity of concurrent object prediction and instruction processing without supplementary contextual information.

The text-based method represents an ablation variant of R$^2$S where segmentation information is omitted, providing only text prompts such as ``\texttt{The question-related objects are chairs, and you may need to pay attention to them}''. Although text-based approach demonstrates modest improvements, our method consistently achieves superior results. This performance advantage can be attributed to our approach's ability to extract precise instance-level visual features through segmentation masks. In contrast, text-based method fails to capture spatial information and demonstrates inherent limitations in differentiating multiple instances of identical categories within a scene (\eg, multiple chairs).

\mypara{The Number of Latent Queries.} We investigate the number of Q-Former latent queries. As demonstrated in \cref{tab:ablation_study_n_latent_queries}, experiments with 16, 32, and 48 latent queries reveal that 32 queries yield optimal performance.

\mypara{Loss Weight.} We conduct an ablation study on the text loss weight $\lambda_{txt}$ using values of 0.1, 0.5, and 1.0. Our experiments demonstrate that setting $\lambda_{txt}$ to 1.0 achieves optimal balance in performance, as evidenced in \cref{tab:ablation_loss_weigths}.

\subsection{Qualitative Results}
As illustrated in \cref{fig:qualitative_results}, we present a comparison between the baseline and R$^2$S. The left portion of the figure compares baseline predictions (upper row) with R$^2$S predictions (lower row) for ScanQA. The lower row shows segmentation results for target-relevant objects. Similarly, the right portion compares baseline and R$^2$S predictions for 3D ReasonSeg. Additional qualitative results are provided in the supplementary material.

%% file: sec/5_conclusions.tex
\section{Limitation}
Due to the use of large language models for data generation instead of human annotation, some noise remains in our dataset even after data cleaning procedures, potentially impacting model performance. We will further optimize our dataset in future work.

\section{Conclusions}
In this paper, we introduce the Relevant Reasoning Segmentation framework, a reasoning-based segmentation framework that enhances spatial reasoning by emulating human reasoning processes through Reasoning Prior Learning and Prior-guided Refinement. Furthermore, we propose 3D ReasonSeg, a reasoning-based segmentation dataset designed to enhance and comprehensively assess the spatial reasoning capabilities in 3D MLLMs. We hope our work provides new insights and directions for exploring the spatial reasoning capabilities in 3D MLLMs.

%% file: X_suppl.tex
\clearpage
\appendix

\section{Data construction details.}
\begin{figure*}[h]
\centering
\resizebox{0.8 \linewidth}{!}{
\includegraphics[]{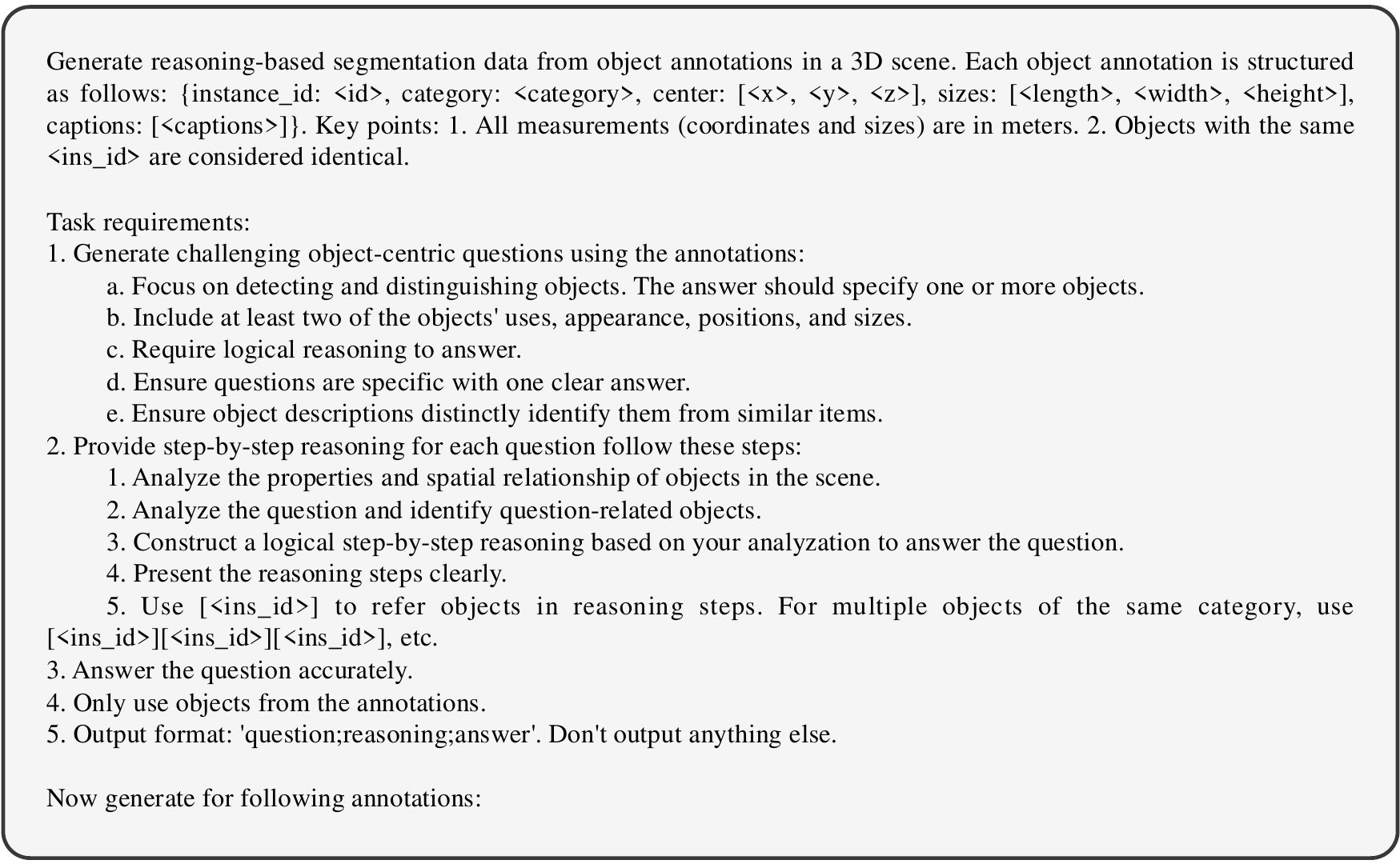}
}
\caption{The prompt for yielding 3D ReasonSeg.}
\label{fig:reasonseg_construction_prompt}
\end{figure*}

\begin{figure*}[h]
\centering
\resizebox{0.8 \linewidth}{!}{
\includegraphics[]{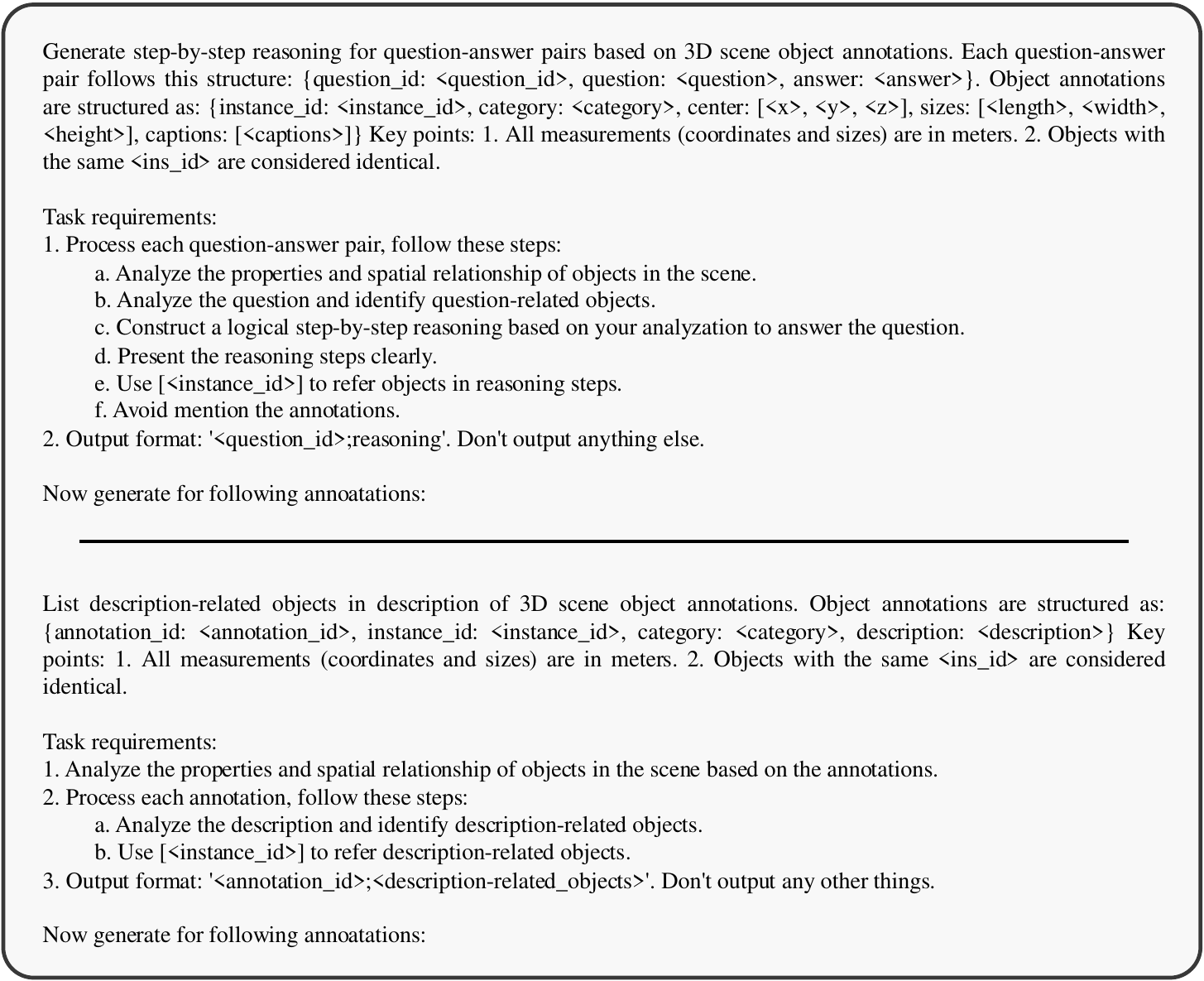}
}
\caption{The prompt for generating target-relevant objects for ScanQA and ScanRefer.}
\label{fig:relevant_objects_construction_prompt}
\end{figure*}

\mypara{Data construction prompts.} The prompt used for data generation is illustrated in \cref{fig:reasonseg_construction_prompt}. We instruct LLama3.1~\cite{dubey2024llama3herdmodels} to create question-reasoning-answer pairs and specify the target-relevant objects in the intermediate reasoning steps. The prompt used to construct target-relevant objects for ScanQA and ScanRefer is depicted in \cref{fig:relevant_objects_construction_prompt}. We instruct LLama3.1 based on the objects' spatial information and annotations from ScanQA and ScanRefer.

\mypara{Rule-based filtering.} In the context of rule-based filtering, we conducted a manual examination of the generated data and formulated filtering strategies. Firstly, we removed data containing phrases such as ``in the right of'' to mitigate potential ambiguity in 3D space. Secondly, for data involving spatial relationships like ``near'', we calculated the distances between the objects and filtered out data with distances surpassing a predefined threshold. Thirdly, in instances of superlative relationships like ``largest'' or ``nearest'', we evaluated object center distances or sizes to validate their adherence to the criteria. Additionally, we excluded data that strayed from the specified prompt structure and identified and eliminated objects created by LLama3.1 based on point cloud instance labels. Furthermore, we pruned data exceeding a designated threshold of related objects and eliminated instances where each reasoning step yielded the same related objects to avoid redundancy. Lastly, we discarded data where reasoning conditions overlapped with reasoning results during inference and where the final step of reasoning conflicted with the answer to the question.

\section{Task formulation details}
We introduce each instruction with the ``\texttt{\textbf{USER:}}'' and format all tasks as auto-regressive generation sequences under the ``\texttt{\textbf{ASSISTANT:}}''.

\textbf{3D Question Answering} tasks the model with responding to questions based on a 3D scene. We utilize a structured format such as: ``\texttt{\textbf{USER:}} Given the 3D scene, please answer the question: \texttt{[QUESTION]}. \texttt{\textbf{ASSISTANT:}} \texttt{[ANSWER]}'', where \texttt{[QUESTION]} and \texttt{[ANSWER]} are sourced from the ScanQA dataset.

\textbf{Scene Description} requires the model to describe the entire scene using natural language. We adopt a format like: ``\texttt{\textbf{USER:}} Describe this scene. \texttt{\textbf{ASSISTANT:}} \texttt{[DESCRIPTION]}'', where \texttt{[DESCRIPTION]} is derived from 3D-LLM.

\textbf{3D Instance Segmentation} task involves segmenting object categories within a scene. Our template is: ``\texttt{\textbf{USER:}} Please segment the {category} in this scene. \texttt{\textbf{ASSISTANT:}} Sure, the segment results are [SEG]...[SEG]'', where the \texttt{[CATEGORY]} is from ScanNet200 and each [SEG] corresponds to an individual object.

\textbf{3D Referring Segmentation} requires the model to segment a target object based on a brief description. Our template is: ``\texttt{\textbf{USER:}} Please segment this object: \texttt{[DESCRIPTION]}. \texttt{\textbf{ASSISTANT:}} Sure, the segment result is [SEG]'', where the \texttt{[DESCRIPTION]} is from ScanRefer and the [SEG] represents the segmentation mask of the target object.

\textbf{3D Reasoning Segmentation} necessitates understanding the visual attributes, functions, and relationships of objects to find the target. Our template is: ``\texttt{\textbf{USER:}} Given the 3D scene, please answer the question with segmentation mask: \texttt{[QUESTION]}. \texttt{\textbf{ASSISTANT:}} \texttt{[ANSWER]}'', where \texttt{[QUESTION]} and \texttt{[ANSWER]} are sourced from the 3D ReasonSeg dataset.

\begin{figure*}[]
\centering
\resizebox{0.9 \linewidth}{!}{\includegraphics{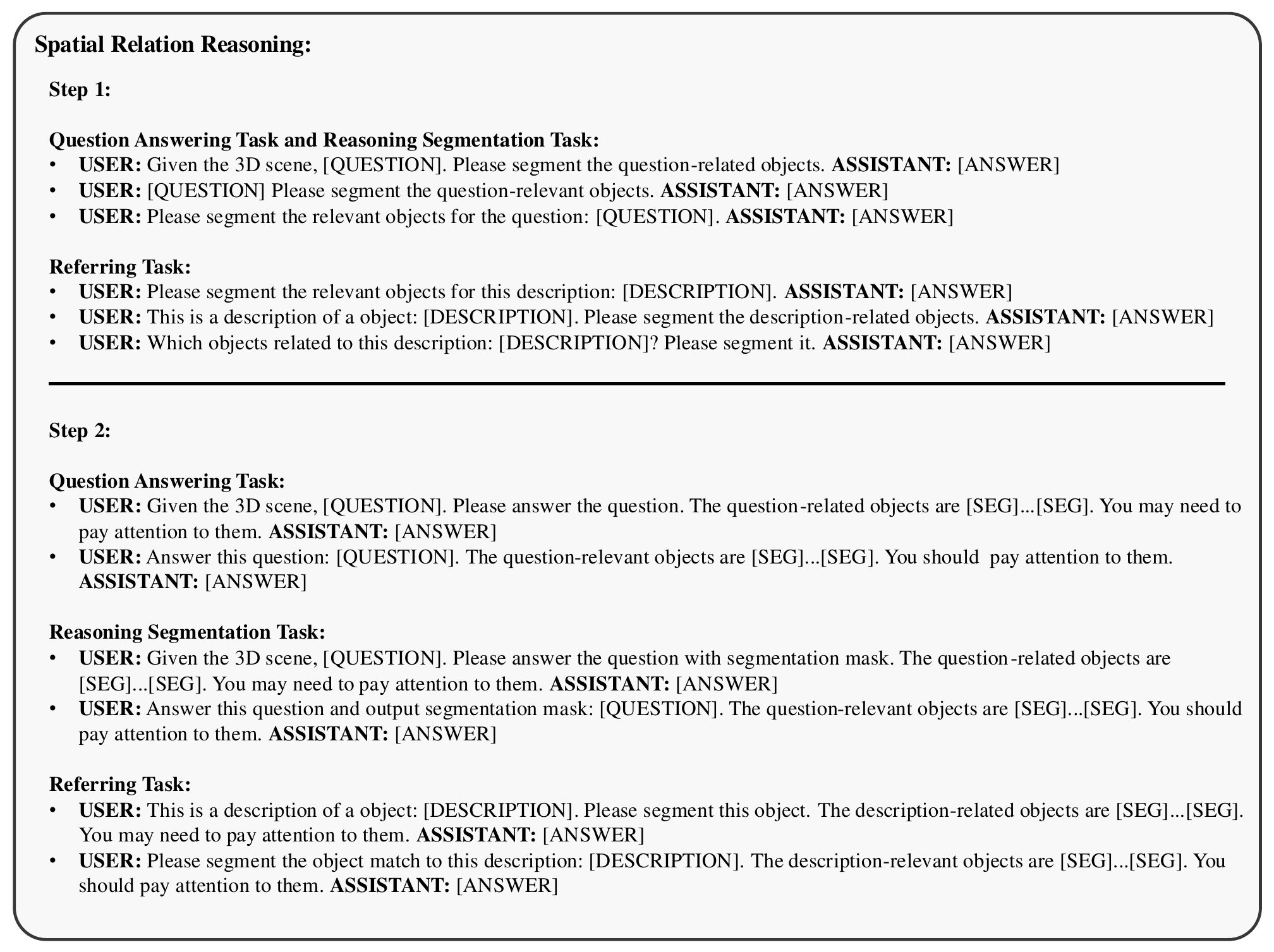}}
\caption{More template for Relevant Reasoning Segmentation.}
\label{fig:reasoning_data_formulation}
\end{figure*}

\mypara{Relevant Reasoning Segmentation Templates.} The details of Relevant Reasoning Segmentation are outlined in the main text, and additional examples for R$^2$S are shown in \cref{fig:reasoning_data_formulation}, covering question answering, reasoning segmentation, and referring.

\begin{table}[h]
\centering
\begin{tabular}{c|ccc}
\toprule
Method & ScanQA & ScanRefer & 3D ReasonSeg
\\ \hline
base & 0.15s & 0.23s & 0.29s
\\
+ R$^2$S & 0.46s & 0.62s & 0.67s
\\
\bottomrule
\end{tabular}
\caption{The efficiency of our proposed methods in ScanQA, ScanRefer and 3D ReasonSeg.}
\label{tab:efficiency_of_methods}
\end{table}

\begin{table*}[h]
\centering
\resizebox{0.9 \linewidth}{!}{
\begin{tabular}{l|l}
\toprule
Task Name & Instruction Template \\ \hline
\multicolumn{1}{l|}{\multirow{2}{*}{3D Reasoning Segmentation}} & \mypara{USER}: Given the 3D scene, please answer the question with segmentation mask: [QUESTION]. \mypara{ASSISTANT}: [ANSWER] \\
& \mypara{USER}: Please answer the question and output corresponding segmentation mask: [QUESTION]. \mypara{ASSISTANT}: [ANSWER] \\
\hline
\multicolumn{1}{l|}{\multirow{2}{*}{3D Question Answering}} & \mypara{USER}: Given the 3D scene, please answer the question: [QUESTION]. \mypara{ASSISTANT}: [ANSWER] \\
& \mypara{USER}: Answer the question based on the 3D scene: [QUESTION]. \mypara{ASSISTANT}: [ANSWER] \\
\hline
\multicolumn{1}{l|}{\multirow{2}{*}{3D Instance Segmentation}} & \mypara{USER}: Given the 3D scene, please segment the [CATEGORY]. \mypara{ASSISTANT}: [ANSWER] \\
& \mypara{USER}: Please segment the [CATEGORY] in this 3D scene. \mypara{ASSISTANT}: [ANSWER] \\
\hline
\multicolumn{1}{l|}{3D Scene Caption} & \mypara{USER}: Please describe the 3D scene. \mypara{ASSISTANT}: [ANSWER] \\
\hline
\multicolumn{1}{l|}{\multirow{2}{*}{3D Referring Segmentation}} & \mypara{USER}: Given the 3D scene, please segment this object: [DESCRIPTION].\mypara{ASSISTANT}: [ANSWER] \\
& \mypara{USER}: Following are descriptions of an object: [DESCRIPTION]. \mypara{ASSISTANT}: [ANSWER] \\
\bottomrule
\end{tabular}
}
\caption{Data formulation for general pre-training.}
\label{tab:general_data_formulation}
\end{table*}

\section{Efficiency of Method.}
As shown in \cref{tab:efficiency_of_methods}, we have evaluated the inference time of our proposed methods on ScanQA~\cite{azuma2022scanqa}, ScanRefer~\cite{chen2020scanrefer}, and 3D ReasonSeg, utilizing a single H20 GPU.

\begin{figure*}[h]
\centering
\resizebox{0.9 \linewidth}{!}{\includegraphics{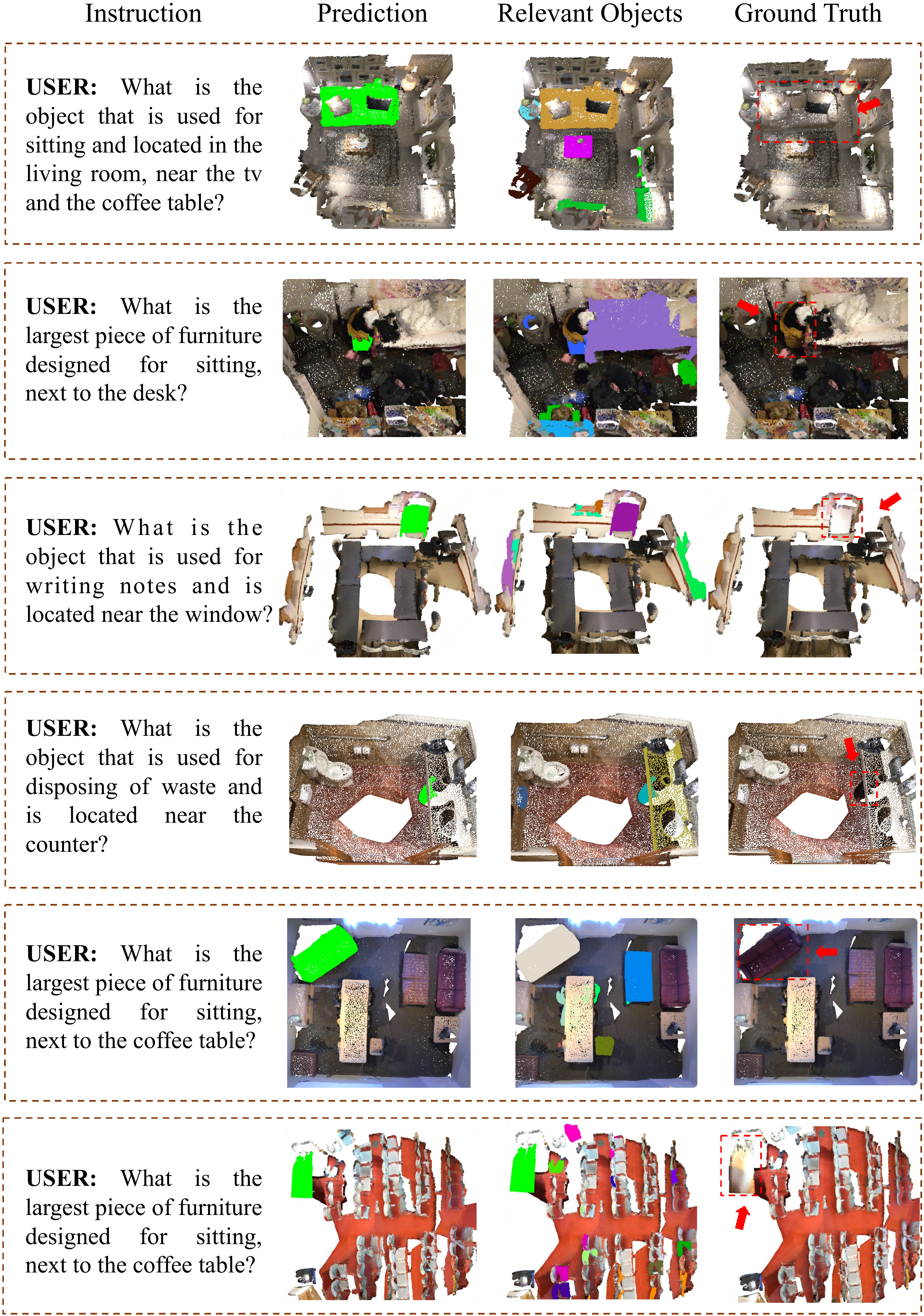}}
\caption{More qualitative results on 3D ReasonSeg.}
\label{fig:more_qualitative_results_3drs}
\end{figure*}

\begin{figure*}[h]
\centering
\resizebox{0.9 \linewidth}{!}{
\includegraphics{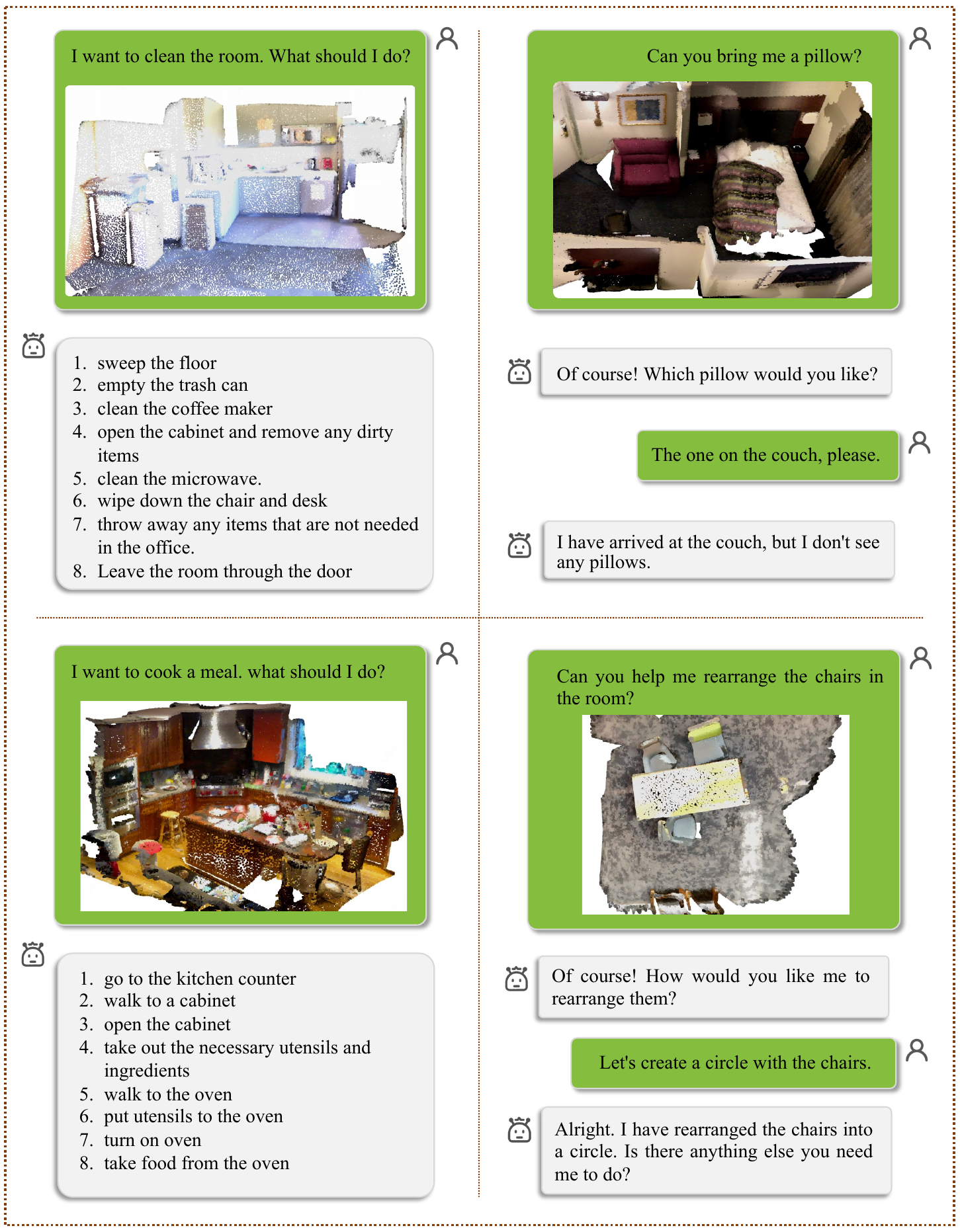}
}
\caption{More qualitative results on scene description and embodied planning.}
\label{fig:more_qualitative_results_3dllm}
\end{figure*}

\section{Additional Qualitative Results}
Additional visual qualitative results of 3D ReasonSeg are shown in \cref{fig:more_qualitative_results_3drs}. Each row displays the user instruction, prediction, target-relevant objects, and ground truth. Further visual qualitative results of dialogue and embodied planning are depicted in \cref{fig:more_qualitative_results_3dllm}.

\section{Additional Implementation Details}

Recent advances in deep learning ~\citep{peng2024oa, tian2020prior, tian2019learning, jiang2021guided, peng2023hierarchical, tian2023learning}, computer vision ~\citep{ning2023boosting, tian2022adaptive, kolodiazhnyi2024oneformer3d, cui2022reslt, tian2022generalized}, and large language models ~\citep{yang2023lisa++, li2023blip, peng2024scalable, yang2024unified, shao2024explore, lai2024step, devlin2018bert, tang2024mind, yang2025visionzip} have significantly advanced the development of perception systems, inspiring our model design.

\mypara{Point Cloud Encoder.} We employ the UNet style SparseConv from OneFormer3D~\cite{kolodiazhnyi2024oneformer3d}, pre-trained on ScanNet. The feature dimension in the Encoder rises from 32 to 160 and then decreases back to 32. The voxel size of the SparseConv is 0.02m. All the parameters of the Encoder are trainable during training.

\mypara{Mask Decoder.} The mask decoder is also derived from the pre-trained OneFormer3D~\cite{kolodiazhnyi2024oneformer3d} model, featuring a 6-layer transformer decoder with a feature dimension of 256. We substitute the default queries with our designated \texttt{[SEG]} token for decoding the segmentation masks.

\mypara{Visual-Language Connector.} We employ QFormer~\cite{li2023blip} as the visual-language connector to condense dense point cloud features into adaptable latent queries. The architecture consists of a 6-layer QFormer with a 768-dimensional feature space, initialized with pre-trained word and positional embeddings from BERT~\cite{devlin2018bert}. The model maintains a vocabulary size of 30,522 and 512 position embeddings.